%% file: main.tex
\documentclass[10pt,twocolumn,letterpaper]{article}

\usepackage{cvpr}              

\usepackage{graphicx}
\usepackage{amsmath}
\usepackage{amssymb}
\usepackage{booktabs}
\usepackage{bbm}
\usepackage[accsupp]{axessibility} %
\usepackage[pagebackref,breaklinks,colorlinks]{hyperref}

\usepackage[capitalize]{cleveref}
\crefname{section}{Sec.}{Secs.}
\Crefname{section}{Section}{Sections}
\Crefname{table}{Table}{Tables}
\crefname{table}{Tab.}{Tabs.}

\def\cvprPaperID{29} %
\def\confName{CVPR}
\def\confYear{2022}

\begin{document}

\title{Lifelong Wandering: A realistic few-shot online continual learning setting}

\author{Mayank Lunayach\thanks{Correspondence to: Mayank Lunayach lunayach@gatech.edu} , James Seale Smith, Zsolt Kira \\
\normalsize
Georgia Institute of Technology
}
\maketitle

\input{latex/sections/abstract}
\input{latex/sections/intro}
\input{latex/sections/related_work}

\input{latex/sections/setting}

\input{latex/sections/baselines}

\input{latex/sections/experiments}

\input{latex/sections/conclusion}

\vspace{-2.5mm}

{\small
\bibliographystyle{ieee_fullname}
\bibliography{egbib}
}

\end{document}

%% file: latex/sections/abstract.tex
\begin{abstract}

Online few-shot learning describes a setting where models are trained and evaluated on a stream of data while learning emerging classes. While prior work in this setting has achieved very promising performance on instance classification when learning from data-streams composed of a single indoor environment, we propose to extend this setting to consider object classification on a series of several indoor environments, which is likely to occur in applications such as robotics. Importantly, our setting, which we refer to as online few-shot continual learning, injects the well-studied issue of catastrophic forgetting into the few-shot online learning paradigm. In this work, we benchmark several existing methods and adapted baselines within our setting, and show there exists a trade-off between catastrophic forgetting and online performance. Our findings motivate the need for future work in this setting, which can achieve better online performance without catastrophic forgetting\footnote{Accepted to CVPR 2022 Workshop on Continual Learning}.
\end{abstract}

%% file: latex/sections/intro.tex
\vspace{-4mm}
\section{Introduction}
Autonomous agents are fed with never-ending data streams and often encounter abrupt changes in the surrounding environments, for example a robot entering a previously unseen building. Learning in this setting is difficult because the data is not identically and independently distributed (iid), an assumption which most deep learning systems rely on. This can specifically lead to a phenomena known as catastrophic forgetting \cite{French1999CatastrophicFI}, where previously accumulated knowledge is replaced with new knowledge. This motivates the \emph{continual learning} \cite{DeLange2019ContinualLA} setting, in which methods are designed to overcome a trade-off between learning new data versus retaining performance  on previously seen data (i.e., mitigating ``forgetting").

In this work, we are interested in the \emph{online} continual learning setting, where data is seen one at a time and discarded after usage. We specifically propose to generalize the online few-shot continual learning setting recently proposed, called RoamingRooms \cite{Ren2021WanderingWA}, that explores the streaming aspect but not catastrophic forgetting. %
The setting uses an indoor imagery dataset to mimic the experience of an agent wandering in the world.
We incorporate distribution shifts using an agent's movements across different buildings and explicitly measure forgetting unlike previous work~\cite{Ren2021WanderingWA}. Using the setting, we then evaluate several existing methods and analyse the online learning performance and forgetting.
We note that while recent works have explored similar settings~\cite{Mai2022OnlineCL}, these works make many unrealistic assumptions (e.g., defined task boundaries, separate training and evaluation phases, etc.) which are typically not applicable for real-world systems.

To summarize, in this work, we: (1) introduce Continual Wandering, a novel naturalistic online continual learning setting; (2) introduce baselines (both prototypical and end-to-end) that would be suited for this challenging setting; and (3) empirically evaluate the baselines and observe a trade-off between online performance and forgetting.

\begin{figure}[t!]
    \centering
    \includegraphics[width=0.45\textwidth]{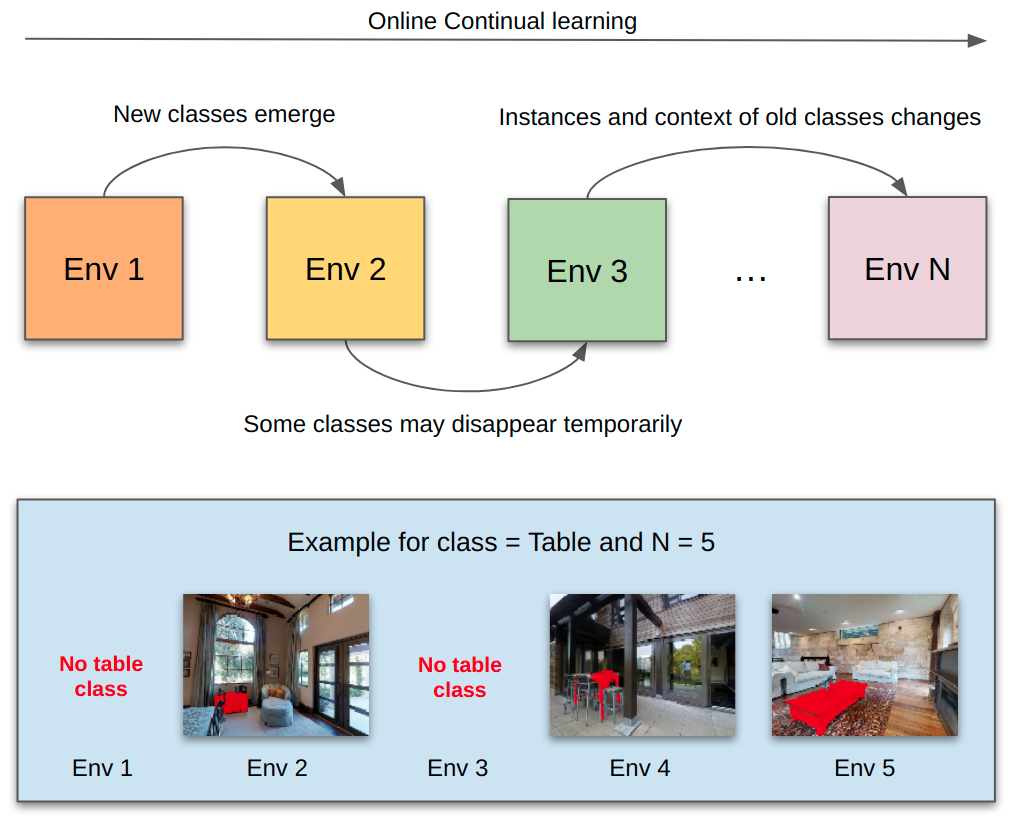}
    \caption{
   Aspects of the proposed setting illustrated for an example using the table class.
   Tables are seen for the first time as a novel class in Env 2. There are no tables in Env 3, but they appear in the fourth and fith environments with different instances and context
    }

    \label{fig:story}
\end{figure}

%% file: latex/sections/related_work.tex
\section{Related Work}\label{sec:related}

Prior works have also pursued more ``realistic" settings for continual learning~\cite{Caccia2020OnlineFA,Goodfellow2014AnEI, Kirkpatrick2017OvercomingCF, Zeno2021TaskAgnosticCL, He2019TaskAC, Harrison2020ContinuousMW, De_Lange_2021_ICCV, koh2021online, Abdelsalam2021IIRCII, Hess2021APW}. 
Our setting is most similar to Wandering Within a World~\cite{Ren2021WanderingWA} (referred to as WW henceforth).
WW proposed an online contextualized few shot learning setting and released the RoamingRooms dataset, based on indoor imagery which mimics experience of an agent wandering within a world. Our work extends the WW setting
with a few key differences. First, the task in WW is to cluster different viewpoints of an object together (i.e., it deals with \emph{instance} classification whereas we deal with \emph{object category classification}).
Second, WW setting does not explicitly capture catastrophic forgetting
 as decrease in its reported accuracy over time could be due to catastrophic forgetting or due to the increasing difficulty because of increasing number of classes. Third, the WW setting is considered
 as an extension of only few-shot learning, and mostly did comparisons against few-shot learning baselines and did not comprehensively benchmark continual learning methods.

%% file: latex/sections/setting.tex
\vspace{-1.5mm}
\section{Setting}\label{sec:setting}

In our setting (visualized in Figure~\ref{fig:story}), an agent navigates a series of environments in an online and continual manner. Let $\boldsymbol{E}_{i=1:N}^{k}$ be the environments in one episode $\mathcal{E}_k$, where $N$ is the total number of environments per episode (we do episodic training and evaluation). Environments are areas from unique buildings (details described later) like kitchen, bedroom, etc.

Environments have a sub-sample of some higher level class distribution, meaning we can have new classes, classes disappearing, or classes persisting from environment to environment (see Figure~\ref{fig:story}). When class persists, the instances and context of the classes across environments is different. For example, say for two different environments, $\boldsymbol{E}_{1}$ and $\boldsymbol{E}_{2}$, a table in $\boldsymbol{E}_{1}$ would be different from a table in $\boldsymbol{E}_{2}$. This difference may arise simply because $\boldsymbol{E}_{1}$ and $\boldsymbol{E}_{2}$ belong to different buildings but same area (say kitchen) or $\boldsymbol{E}_{1}$ and $\boldsymbol{E}_{2}$ belong to different buildings and different areas (say kitchen and living room). This is how natural distribution shifts emerge across environments. This has been illustrated in Figure~\ref{fig:story}. Further, most progress in Online Continual learning has happened on synthetic object centric benchmarks like Omniglot\cite{Lake2015HumanlevelCL}, MNIST\cite{lecun1998gradient}, Tiered-ImageNet\cite{Ren2018MetaLearningFS}, CIFAR\cite{Krizhevsky2009LearningML}, CORe50\cite{Lomonaco2017CORe50AN}, etc. Our setting uses RoamingRooms \cite{Ren2021WanderingWA} that has occlusions in objects and a natural distribution of viewpoints in terms of scale, orientation, etc.

Each environment $\boldsymbol{E}_{i}$ has $T$ frames (captured from a viewpoint of a wandering agent) sampled from one of the Matterport3D\cite{Chang2017Matterport3DLF} \textit{real} buildings through a two-stage process~\cite{Ren2021WanderingWA} using HabitatSim\cite{Savva2019HabitatAP} and MatterSim\cite{Anderson2018VisionandLanguageNI}. 
90 Matterport3D\cite{Chang2017Matterport3DLF} buildings are splitted into 60 for training, 10 for validation and 20 for testing. Each training/evaluation episode $\mathcal{E}_k$ is concatenation of the $N$ environments, i.e. $\mathcal{E}_k = [\boldsymbol{E}_{1}^{k},~\boldsymbol{E}_{1}^{k},~ ...~\boldsymbol{E}_{N}^{k}]$. For training, we sample $N$ environments from 60, thus we have more than ${60 \choose N}$ unique training episodes ($\approx$ half million for typical $N = 4$) available. The number is more than ${60 \choose N}$ because each environment has typically more than one area (like kitchen, bathroom, etc.) to sample from.

Similar to how an agent would explore the real world, our setting is semi-supervised in that we include learning from both labeled and unlabeled data examples. Further, to simulate the real life scenario where an agent would have to do learning and with one sample at time, we only allow online batch size = 1 for our setting, that is, the model updates with one sample at a time. 

Episode $\mathcal{E} = \{x_{t=1:N \times T}, y_{t=1:N \times T}\}$ ($k$ from $\mathcal{E}_{k}$ dropped for simplifying the notation) where $x_t$ is a 4-channel input corresponding to $t_{ \text{th}}$ frame where first three channels are from  the frame's RGB image and the last channel corresponds to binary mask segmenting out the current instance from the background, $y_t$ is a class label if the instance is labelled, $N$ is the number of environments in one episode, and $T$ is the number of frames in one environment. Now we define evaluation protocol for our setting.

\subsection{Evaluation}
As done in \cite{Caccia2020OnlineFA} and \cite{Ren2021WanderingWA}, there is no separate training and evaluation phase.  Because in real life, we learn and remember concepts at the same time. We incorporate this by using an episode training framework where meta-training happens on training set and then later the performance is evaluated on validation/test set. We calculate an agent's online performance as it sees each training example and measure its forgetting by revisiting the prior training examples and measuring the agent's performance degradation on these examples. Specifically, we calculate average online accuracy and forgetting in our setting as

\begin{itemize}
    \item \textbf{Average Online Accuracy}: Let $O_{i}$ be the online accuracy for the environment $i$. We define Average Online Accuracy, $O_avg$,
    
    \begin{equation}
        O_{avg} = \frac{\sum_{i=1}^{N} O_i}{N}
    \end{equation}
    
    where $N$ is the no. of environments in one episode and $O_{i}$ is defined as follows,
    
    \begin{equation}
        O_{i} = \frac{\sum_{k=1}^{K} O_{i}^{k}}{K}
    \end{equation}
    where,
    \begin{equation}
        O_{i}^{k} = \frac{\sum_{t=T'}^{T' + T} \mathbbm{1}(y_{t} = \hat{y}_{t}) \mathbbm{1}(x_{t} \in seen\_class) }{\sum_{t=T'}^{T' + T} \mathbbm{1}(x_{t} \in seen\_class)}
    \end{equation}
    where $x, y$ are as defined in the section above, $\hat{y_t}$ is the model prediction after the model has seen $\{x_{t=1:t-1}, y_{t=1:t-1}\}$, $T$ is the number of frames in one environment, $T'$ is the number of frames seen before environment $i$, $seen\_class$ is the set of classes seen till the $t-1$ timestamp and $K$ is the number of episodes. $O_{i}$ is averaged across the episodes and $O_{avg}$ is averaged across the environments.

    \item \textbf{Average Forgetting}: Let $C_{i,j}$ be the average accuracy on environment $j$ after seeing environment $i$. Forgetting of environment $j$ after seeing environment $i$, $FFF_{i, j}$ defined as
    
    \begin{equation}
        FFF_{i,j} = C_{j,j} - C_{i,j}
    \end{equation}
    
    Average forgetting after seeing environment $i$, $FF_{i}$ defined as
    \begin{equation}
        FF_{i} = \frac{\sum_{j=1}^{i-1} FFF_{i,j}}{i-1} 
    \end{equation}
    
    We report the average forgetting across all environments $F_{N}$, defined as
    
    \begin{equation}
        F_{avg} = \frac{\sum_{i=1}^{N} FF_{i}}{N} 
    \end{equation}
    Now we discuss the baselines evaluated on the setting.
    
\end{itemize}

%% file: latex/sections/baselines.tex
\vspace{-5mm}
\section{Baselines}
As discussed in the Section~\ref{sec:related}, continual learning approaches can be divided into regularisation based, replay based and meta learning based. LwF\cite{Li2016LearningWF}, a regularisation based approach, has known to work well for removing forgetting for online continual learning \cite{Mai2022OnlineCL}. OML\cite{Javed2019MetaLearningRF} is an established meta-learning based objective for mitigating catastrophic forgetting and accelerating online learning. Using replay based approaches makes the setting unrealistic (especially with issues such as storage and privacy), and thus we don't evaluate on such a baseline. CPM\cite{Ren2021WanderingWA} was proposed specifically for the dataset we base our setting on. Thus, we adapt LwF, OML, CPM and its simpler version Online Averaging for our setting.

\noindent \textbf{Contextual Prototypical Memory (CPM)~\cite{Ren2021WanderingWA}}: This prototype-based approach uses a RNN to model spatiotemporal context for online contextualized few-shot learning.

\noindent \textbf{Online Averaged Prototypes (OAP)}: Here, we extend CPM but rather than using  spatiotemporal context for weighting the prototype updates, we do simple online averaging as follows: For a class $c$, let the labelled instances count at time $t-1$ be $count_{c, t-1}$, its class prototype be $P_{c, t-1}$ and the agent sees a new labelled instance feature of class $c$, $f_{c, t}$ at time $t$, then  

\begin{equation}
    count_{c, t} = count_{c, t-1} + 1
\end{equation}    
\begin{equation}
    P_{c, t} = \frac{f_{c, t} + count_{c, t-1} \times P_{c, t-1}}{count_{c, t}}
\end{equation}

\begin{figure}[t]
    \centering
    \includegraphics[width=0.45\textwidth]{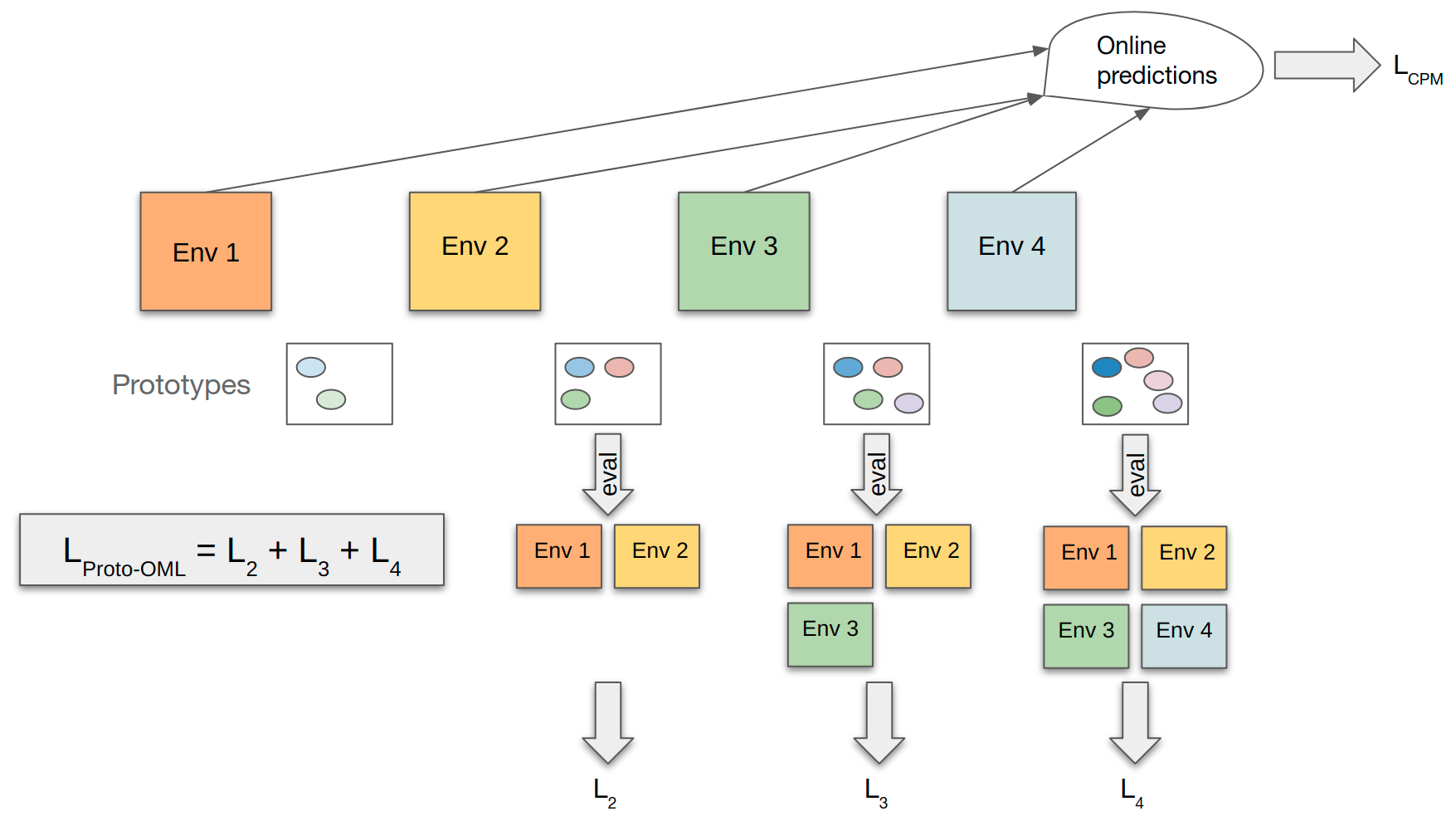}
    \caption{Visualization of our adapted proto-OML baseline
    }
    \label{fig:meta-continual}
\end{figure}
\begin{figure*}[t]
    \centering
    \begin{subfigure}{0.32\textwidth}
        \centering
        \includegraphics[width=\textwidth]{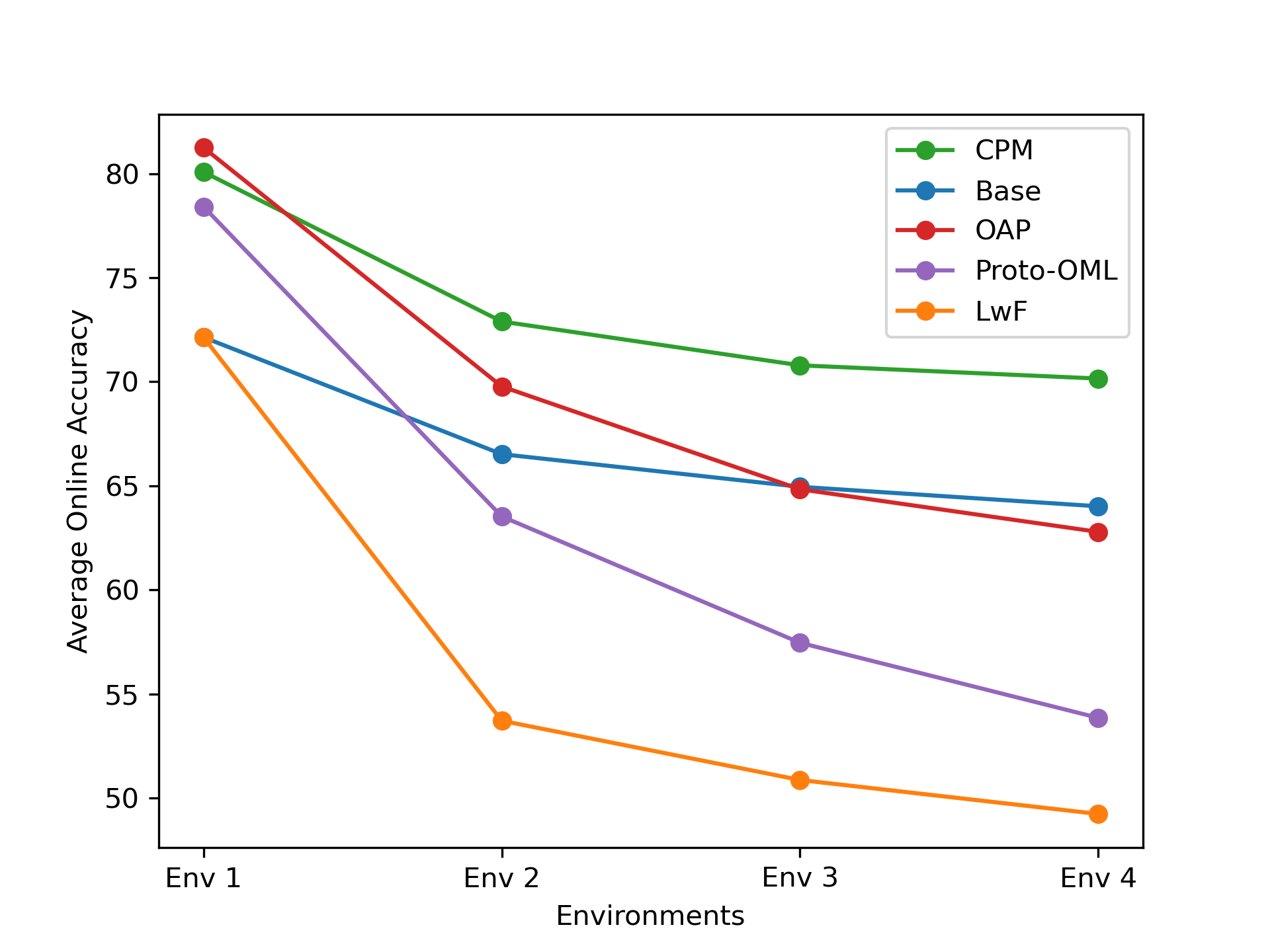}
        \caption{Average Online Accuracy vs Environments
        }
        \label{fig:AOA}
    \end{subfigure}
    \begin{subfigure}{0.32\textwidth}
        \centering
        \includegraphics[width=\textwidth]{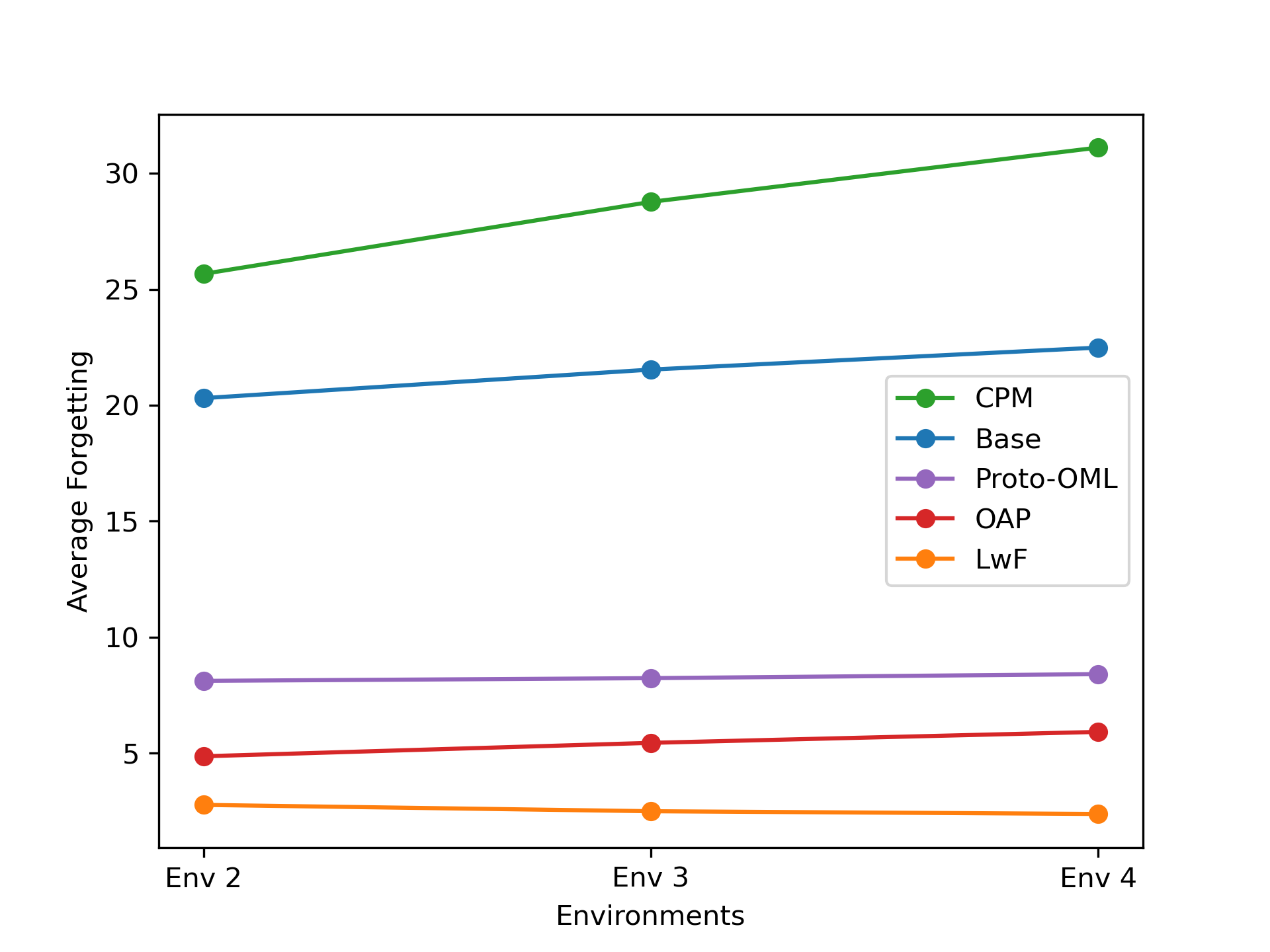}
        \caption{Average Forgetting vs Environments
        }
        \label{fig:AF}
\end{subfigure}
    \begin{subfigure}{0.32\textwidth}
        \centering
        \includegraphics[width=\textwidth]{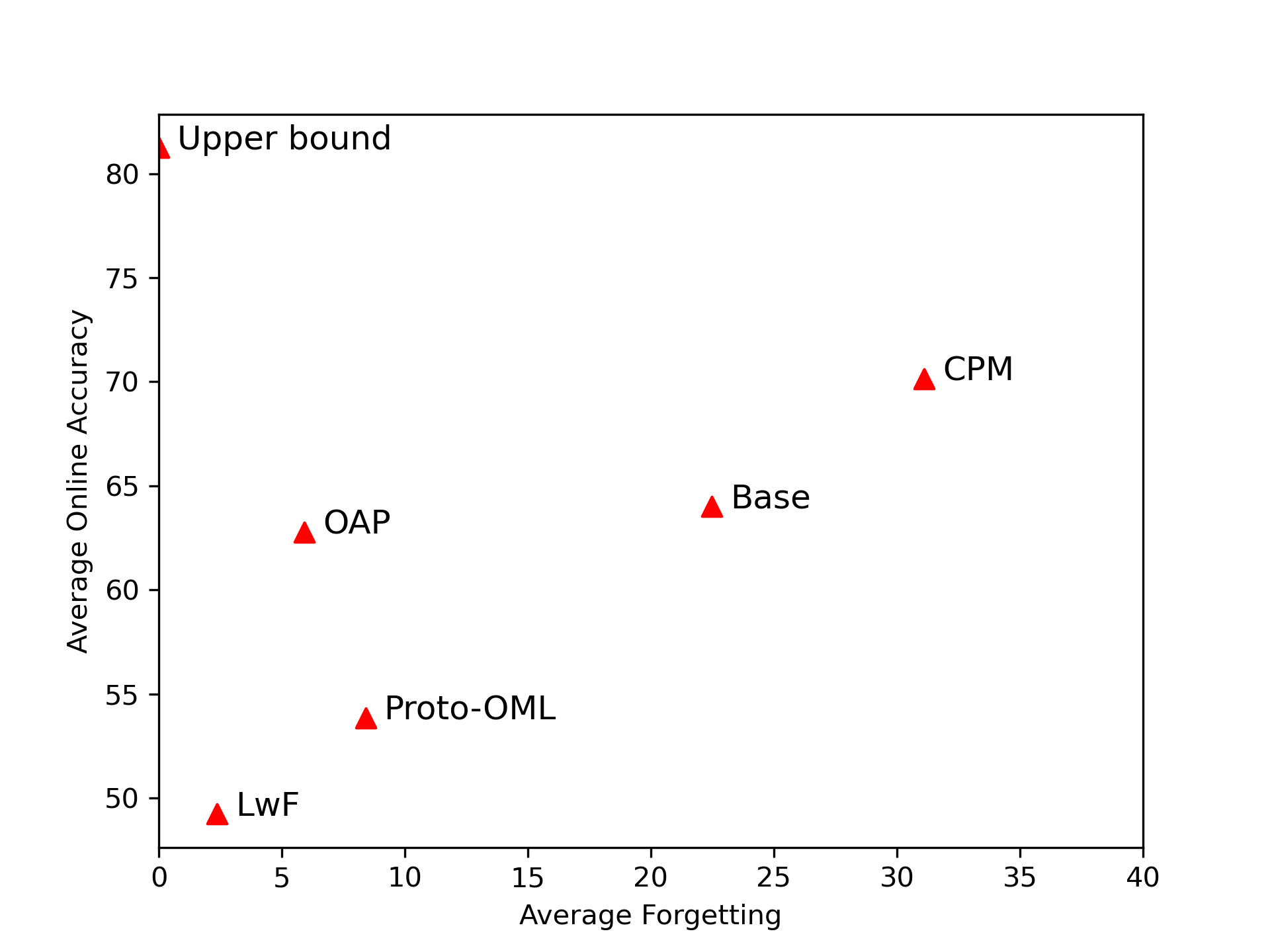}
        \caption{Average Online Accuracy vs Forgetting for the last environment
        }
        \label{fig:LastAccVsForget}
    \end{subfigure}
    \vspace{-0.25cm}
    \caption{A closer look at average online accuracy and average forgetting}
    \vspace{-0.25cm}
\end{figure*}

\noindent \textbf{Proto-OML}: For this baseline, we extend Online-aware Meta Learning (OML)\cite{Javed2019MetaLearningRF} to be effective for our setting. OML proposes a meta learning based objective to learn not to forget the seen classes. Our extension is prototype based as shown in the Figure~\ref{fig:meta-continual}.  
Specifically, at the end of each environment, we calculate loss on all the seen environments. That is, say the agent has just seen the environment $i$, then $L_{i}$ is defined as follows,

\begin{equation}
L_{i} = \frac{\sum_{t=1}^{T'} CrossEntropy(y_{t}, \hat{y}_{t}) \mathbbm{1}(x_{t} \text{is labelled}) }{\sum_{t=1}^{T'} \mathbbm{1}(x_{t} \text{is labelled})}
\end{equation}

\begin{equation}
T' = i \times T
\end{equation}
where $T$ is the number of frames in one environment, $N$ is the number of environments in one episode, and predictions $\hat{y}_{1:T'}$ are outputs of an additional forward pass on $T'$ frames after the agent has seen the environment $i$.

Proto-OML loss $L_{Proto-OML}$ can be defined as following,
\begin{equation}
    L_{Proto-OML} = \sum_{i=2}^{N} L_{i}
\end{equation}
and the total loss $L_{total}$ becomes,
\begin{equation}
    L_{total} = L_{CPM} + \lambda L_{Proto-OML}  
\end{equation}
where $L_{CPM}$ is the same loss as used in the first baseline and $\lambda$ is a hyper-parameter.

\noindent \textbf{LwF}: 
LwF~\cite{Li2016LearningWF} uses knowledge distillation to preserve knowledge from the previously seen tasks. 
In order to fairly adapt LwF to our setting, the model is trained offline on the entire training dataset and then the weights of the final linear classification layer are re-initialized for every episode with the weights of the feature extractor kept frozen.

\noindent \textbf{Base}: In this baseline, one linear layer (classifier) is added after the feature extractor. Only the classifier is trained with the feature extractor kept frozen (which was trained during the offline pre-training step).

\noindent \textbf{Upper bound}: In this baseline, we use OAP but re-initialize the prototypes at the start of every environment. This ensures that the online performance is not affected by accumulation of prototypes of old instances and gives us an upper bound of the online performance.

%% file: latex/sections/experiments.tex
\section{Experiments}
\noindent \textbf{Details}: 
Training and evaluation happens using episodes generated from the RoamingRooms \cite{Ren2021WanderingWA} dataset. 
During evaluation, the feature extractor in each method is frozen and only prototypes (for prototype based methods) or linear classification layer (for LwF and Base) are updated. All methods use the ResNet-12 backbone as their feature extractor. Further, prototype based methods (CPM, OAP and Proto-OML) use 512-dimensional prototypes to store the classes information and use cosine similarity between the extracted features and the prototypes to generate the class prediction scores. To keep the comparisons fair, base and LwF use one linear layer with hidden dimension = 512 after the backbone to generate the prediction scores. 

The number of environments ($N$) per episode is 4, number of frames per environment ($T$) is 100, and the fraction of labelled data is kept 0.4. Hyperparameters are optimized on the validation set. Batch size is 1 for all the methods. Prototypical methods are trained with 4000 episodes (thus 4000 iterations) with learning rate = 1e-4 which decays to 1e-5 and 1e-6 after 2000 and 3000 steps respectively using Adam \cite{Kingma2015AdamAM} optimizer. LwF and base are first pre-trained offline using a dataset created by sampling frames uniformly across all the training episodes combined. For offline training, out of 60 training episodes, 48 are used for training ($\approx 100000$ frames) and 12 are used for validation ($\approx 25000$ frames). LwF and Base are evaluated (after initializing their feature extractor weights with pre-trained's) using learning rate = 0.27 and SGD \cite{Robbins2007ASA} optimizer.

\begin{table}[t]
\begin{center}
\caption{Online Average Accuracy ($O_{avg}$) and Average Forgetting ($F_{avg}$) on the Test set}
\begin{tabular}{|c | c | c|} 
 \hline
 Method & $O_{avg}$ ($\uparrow$) & $F_{avg}$ ($\downarrow$) \\ [0.5ex] 
 \hline
 Upper bound & $81.25 \pm 9.66$ & - \\
 \hline
 Base & $66.76 \pm 5.85$ & $22.48 \pm 8.27$ \\
 \hline
 LwF \cite{Li2016LearningWF} & $56.03 \pm 6.79$ & $2.38 \pm 4.82$ \\
 \hline
 Proto-OML & $62.75 \pm 6.06$ & $8.41 \pm 5.04$ \\
 \hline
  CPM\cite{Ren2021WanderingWA} & $73.26 \pm 5.11$ & $31.11 \pm 7.65$ \\
  \hline
  OAP & $69.23 \pm 5.57$ & $5.91 \pm 3.39$ \\
    \hline
\end{tabular}
\label{tab:results}
\end{center}
\vspace{-0.6cm}
\end{table}

\noindent \textbf{Discussion}: Our results are summarized in Table \ref{tab:results}. We see that $O_{avg}$ decreases as the agent sees more environments for all the methods. The decline is minimum for CPM (Figure~\ref{fig:AOA}) which is a spatiotemporal context method suggesting keeping track of context helps the agent weigh recent information more. But this comes at a cost as forgetting of CPM is the highest (Figure ~\ref{fig:AF}). Surprisingly, OAP (a simpler version of CPM) is little inferior to CPM it online performance but has far less forgetting. This suggests equally weighting the inputs is more effective than context based weighting for our setting. LwF, an established rehearsal free continual learning method has negligible forgetting but this comes with very poor online performance. This hints that LwF is reducing forgetting at the cost of learning new information (the forgetting vs plasticity dilemma). \textbf{Importantly, our findings suggest that methods designed for the ``extremes" (i.e., online accuracy \emph{or} forgetting) perform poorly when evaluated from our unified perspective (i.e., online accuracy \emph{and} forgetting).} Simple methods such OAP seem to work best (Figure~\ref{fig:LastAccVsForget}) but there still remains a large gap for future work.

%% file: latex/sections/conclusion.tex
\section{Conclusion}
\vspace{-2mm}
In this work, we benchmark existing strategies in a realistic setting for few-shot online continual learning. We found that methods which have good online performance have high forgetting and vice versa. Simple methods such as online averaging seem to be most promising but there is still much room for improvement. We hope this setting and the empirical evaluation in this article drives more progress in online lifelong learning.